\documentclass{article}

\usepackage{xcolor}
\usepackage{arxiv}

\usepackage[utf8]{inputenc} 
\usepackage[T1]{fontenc}    
\usepackage{hyperref}       
\usepackage{url}            
\usepackage{booktabs}       
\usepackage{amsfonts}       
\usepackage{nicefrac}       
\usepackage{microtype}      

\usepackage{graphicx}
\usepackage{doi}
\usepackage{amsmath}

\usepackage[numbers]{natbib}
\usepackage{subcaption}
\usepackage[nameinlink]{cleveref}
\usepackage{float}

\newcommand{\cacti}{$\textrm{CACTI}$}
\newcommand{\cactif}{$\textrm{CACTI}_\textrm{F}$}

\title{Style Transfer with Diffusion Models for Synthetic-to-Real Domain Adaptation}

\author{\href{https://orcid.org/0009-0001-8659-6006}{\includegraphics[scale=0.06]{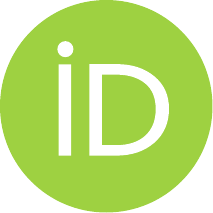}\hspace{1mm}Estelle Chigot}\\
	ISAE-Supaero, University of Toulouse, France\\
	Airbus, France\\
	\texttt{estelle.chigot2@isae.fr} \\
	\And
	\href{https://orcid.org/0000-0003-2414-0051}{\includegraphics[scale=0.06]{images/orcid.pdf}\hspace{1mm}Dennis G. Wilson} \\
	ISAE-Supaero, University of Toulouse, France\\
	\texttt{dennis.wilson@isae.fr} \\
	\AND
	\href{https://orcid.org/0009-0008-7443-7340}{\includegraphics[scale=0.06]{images/orcid.pdf}\hspace{1mm}Meriem Ghrib} \\
	Airbus, France\\
	\texttt{meriem.ghrib@airbus.com} \\
    \hspace{6.6cm}
	\And
	\href{https://orcid.org/0000-0002-9680-4227}{\includegraphics[scale=0.06]{images/orcid.pdf}\hspace{1mm}Thomas Oberlin} \\
	ISAE-Supaero, University of Toulouse, France\\
	\texttt{thomas.oberlin@isae.fr} \\
}

\hypersetup{
pdftitle={Style Transfer with Diffusion Models for Synthetic-to-Real Domain Adaptation},
pdfsubject={cs.LG, cs.CV},
pdfauthor={Estelle Chigot, Dennis G. Wilson, Meriem Ghrib, Thomas Oberlin},
pdfkeywords={Semantic segmentation, Synthetic data, Domain adaptation, Style transfer, Diffusion model},
}

\renewcommand{\shorttitle}{Style Transfer with Diffusion Models for Synthetic-to-Real Domain Adaptation}  

\begin{document}

\textit{Published in Computer Vision and Image Understanding, September 2025 (CVIU 2025)}

\let\WriteBookmarks\relax
\def\floatpagepagefraction{1}
\def\textpagefraction{.001}

\date{}
\maketitle

\begin{abstract}
Semantic segmentation models trained on synthetic data often perform poorly on real-world images due to domain gaps, particularly in adverse conditions where labeled data is scarce.
Yet, recent foundation models enable to generate realistic images without any training. This paper proposes to leverage such diffusion models to improve the performance of vision models when learned on synthetic data.
We introduce two novel techniques for semantically consistent style transfer using diffusion models: Class-wise Adaptive Instance Normalization and Cross-Attention (\cacti) and its extension with selective attention Filtering (\cactif).
\cacti\ applies statistical normalization selectively based on semantic classes, while \cactif\ further filters cross-attention maps based on feature similarity, preventing artifacts in regions with weak cross-attention correspondences.
Our methods transfer style characteristics while preserving semantic boundaries and structural coherence, unlike approaches that apply global transformations or generate content without constraints.
Experiments using GTA5 as source and Cityscapes/ACDC as target domains show that our approach produces higher quality images with lower FID scores and better content preservation.
Our work demonstrates that class-aware diffusion-based style transfer effectively bridges the synthetic-to-real domain gap even with minimal target domain data, advancing robust perception systems for challenging real-world applications. The source code is available at: \texttt{\textcolor{magenta}{\url{https://github.com/echigot/cactif}}}.
\end{abstract}


\section{Introduction}\label{intro}

Semantic segmentation represents a fundamental computer vision task that assigns a class label to each pixel in an image, enabling detailed scene understanding.
This capability forms a critical foundation for numerous applications including autonomous driving, robotics, and medical image analysis.
While deep learning approaches have achieved remarkable results in this domain, their success depends heavily on the availability of large, accurately labeled datasets.

The creation of such datasets requires extensive manual annotation, a process that is both time-consuming and costly.
For example, annotating a single image in the Cityscapes dataset of urban views from a car takes approximately 90 minutes of human effort \cite{cordts2016Cityscapes}.
Synthetic data generated from graphic engines offer a compelling alternative, providing automatically labeled images at scale.
However, models trained exclusively on synthetic data typically perform poorly when evaluated on real-world images due to the domain gap—differences in visual characteristics between synthetic and real domains including lighting conditions, texture details, and object appearances.

This synthetic-to-real domain gap has motivated extensive research in domain adaptation techniques, which aim to transfer knowledge from a labeled source domain (synthetic) to an unlabeled or sparsely labeled target domain (real).
Existing approaches typically fall into three categories: feature-level adaptation that aligns representations in the model's latent space \cite{hoffman2016fcns,hoyer2022daformer}, output-level adaptation that focuses on prediction consistency \cite{chattopadhyay2024augcal}, and input-level adaptation that transforms source images to resemble the target domain \cite{benigmim2023datum,jia2024dginstyle,wang_domain_2024}.

Among input-level methods, style transfer is a promising approach.
By transferring the visual characteristics of target domain images to source domain content, these techniques can generate training data that maintains the accurate labels of the source while exhibiting the appearance of the target.
Traditional style transfer methods have focused primarily on artistic applications, with limited exploration of their potential for domain adaptation in semantic segmentation tasks \cite{alaluf2024cross,go2024eye,hertz2024style}.

Recent advances in diffusion models have greatly improved image generation capabilities, demonstrating unprecedented quality and flexibility in content creation.
These models, which generate images through an iterative denoising process, offer powerful new mechanisms for style transfer that could potentially address the synthetic-to-real domain gap more effectively than previous approaches.
However, applying diffusion-based style transfer specifically for domain adaptation presents unique challenges that remain largely unexplored.

In this paper, we investigate diffusion-based style transfer for synthetic-to-real domain adaptation in semantic segmentation, with a particular focus on few-shot scenarios where only limited target domain data is available.
We identify two key limitations in existing approaches: (1) standard style transfer methods apply global transformations that ignore semantic class boundaries, leading to unrealistic appearances within object categories, and (2) direct application of cross-attention mechanisms can create artifacts when structural correspondence between domains is weak.

To address these challenges, we propose two novel techniques:
First, \textbf{C}lass-wise \textbf{A}daptive Instance Normalization and \textbf{C}ross-atten\textbf{TI}on (\cacti) applies statistical normalization selectively based on semantic classes, ensuring that style characteristics are transferred appropriately within semantic boundaries.
Second, \cacti\ with Selective Attention \textbf{F}iltering (\cactif) extends this approach by selectively applying cross-attention based on feature similarity, preventing artifacts in regions where direct style transfer might produce inconsistencies.

Our experiments demonstrate that these techniques not only produce visually coherent images but also improve semantic segmentation performance when used for domain adaptation.
Particularly noteworthy is their effectiveness in challenging adverse weather conditions such as fog, rain, snow, and nighttime, where the domain gap is especially pronounced.
When combined with state-of-the-art segmentation architectures, our approach improves over existing methods, highlighting the potential of semantically-aware diffusion-based style transfer for addressing the synthetic-to-real domain gap.

\section{Related work}\label{related-work}

\subsection{Style Transfer with Diffusion Models}

Diffusion models generate images through an iterative denoising process that progressively refines Gaussian noise into coherent visual outputs. Recent advancements such as DDPM \cite{ho2020denoising} and DALL·E 2 \cite{ramesh2022hierarchical} have demonstrated the ability to produce images of high quality and diversity. Stable Diffusion (SD) \cite{rombach2022high} further improves efficiency by performing the denoising process in the latent space of a Variational Autoencoder (VAE), using a UNet-based architecture. This design significantly reduces computational cost while maintaining high fidelity in generated images. Building on this foundation, ControlNet \cite{zhang2023adding} introduces a conditioning mechanism that enables control over image generation using inputs such as human poses or segmentation maps. By incorporating zero-convolution layers during training, ControlNet aligns conditional inputs with the denoising process, allowing for fine-grained manipulation of image attributes.

A subclass of image generation methods focuses on transferring style characteristics from a reference image to a target content image—a process known as style transfer. Early approaches, such as the method proposed by Gatys et al. \cite{gatys2016image}, utilized Convolutional Neural Networks (CNNs) to separately extract content and style features by analyzing activations at different convolutional layers. To improve flexibility and efficiency, Adaptive Instance Normalization (AdaIN) \cite{huang2017arbitrary} was introduced. AdaIN aligns the channel-wise mean and variance of the style features with those of the content image in the latent space of a CNN, enabling arbitrary style transfer in real-time.
In parallel, Generative Adversarial Networks (GANs) \cite{heusel2017fid} have also been widely adopted for style transfer tasks \cite{zhu2017unpaired,chen2019gatedgan,lin2021gan-daynight}, at the expense of training stability and with a risk of mode collapse.

Developments in diffusion models have made them suitable to perform style transfer tasks. Several methods leverage text-based algorithms to stylize images by prompting a pretrained diffusion model without the need for finetuning. These approaches use textual descriptions to guide the generation process and achieve style transfer through prompt engineering and latent manipulation \cite{zhang2023inversion, qi2024deadiff, jeong2024visual, hertz2024style}.
Beyond text-based guidance, other techniques enable style transfer by directly blending features from two input images during inference. Cross-image attention \cite{alaluf2024cross}, Z* \cite{deng2024z} and StyleID \cite{chung_style_2024} use KV injection in the self-attention layers of Stable-Diffusion's UNet. Eye-for-an-eye \cite{go2024eye} rearranges the features in those self-attention layers following a computed semantic correspondence.
Additionally, some recent works incorporate Low-Rank Adaptation (LoRA) \cite{hu2022lora}, a parameter-efficient finetuning technique, to enable more flexible style-content fusion. These methods, such as B-LoRA \cite{frenkel2024implicit} and ZipLoRA \cite{shah2024ziplora}, fine-tune lightweight layers to blend styles from multiple reference images.

Those methods are generally used to generate artistic images, and only few of them are actually employed as data generation pipelines for other vision tasks. However, due to their large training datasets and generalization capabilities, diffusion based style transfer methods seem appropriate to bridge the synthetic to real domain gap. 

\subsection{Synthetic to Real Domain Adaptation}

Synthetic to real, or Sim-to-real, domain adaptation refers to the attempt of closing the gap between training on simulated data and testing on real-life data. This topic is particularly relevant when data is scarse, expensive to annotate and scenes are complex such as in robotics or autonomous driving. 
Classical domain adaptation techniques typically rely on access to large, labeled datasets in both source and target domains. 
However, more recent approaches focus on one-shot or zero-shot adaptation, using limited to no target domain data. 
In the context of semantic segmentation, DAFormer \cite{hoyer2022daformer} introduces a transformer-based architecture that leverages strong feature encoding and pseudo-label refinement, achieving state-of-the-art performance on synthetic-to-real benchmarks. Extending this work, HRDA \cite{hoyer2022hrda} enhances performance by incorporating high-resolution feature fusion at multiple scales, allowing the model to better capture fine-grained details and spatial consistency across domains.
PØDA \cite{fahes2023poda} uses CLIP to integrate prompts embeddings to align the source domain and the target domain, without having access to target annotations. 

Leveraging the generation capacity of diffusion models, some methods have integrated those into domain adaptation pipelines. SGG \cite{peng2023diffusion} uses stable diffusion to generation intermediate domains during the segmentation training mechanism. Gong et al.~\cite{gong_prompting_2023} plug a segmentation head onto a diffusion backbone, making use of the disantangled representations of style and content learned by diffusion models. 

DGInStyle \cite{jia2024dginstyle} finetunes Dreambooth \cite{ruiz2023dreambooth} on the source domain, then train ControlNet to generate images conditioned by segmentation maps. Finally, they apply a style swap operation, switching the source specific UNet to a generalist pretrained one. In this manner they only retain the segmentation conditioning without style leaking from the source dataset during the ControlNet training. 
DATUM \cite{benigmim2023datum} also uses Dreambooth to finetune stable diffusion on only one target image for a few iterations. Then, they prompt the model to generate one target object in the style of the target image. 
Those two generation methods can be used to generate datasets tailored to the domain adaptation use-case, in a zero-shot setting for DGInStyle or one-shot setting for DATUM.

Diffusion based style transfer has not yet been used commonly to generate datasets for domain adaptation.
DoGE \cite{wang_domain_2024} computes the mean difference of CLIP embeddings pairs between the source and target dataset (Domain Gap Embedding). Then, it adds this Domain Gap Embedding to the latent representation of a source image, before reconstructing the image with Stable UnCLIP. This method can use ControlNet, allowing the model to generate datasets conditioned on semantic segmentation maps.
In this work, we propose to use Cross-image attention \cite{alaluf2024cross} as our style-transfer method to generate datasets for its zero-shot generation capability. 

\section{Methodology}

We introduce two novel techniques to address these limitations: Class-wise Adaptive Instance Normalization with Attention (\cacti), possibly combined with Selective Attention Filtering (\cactif).
 \begin{figure*}
     \centering
     \includegraphics[width=1\linewidth]{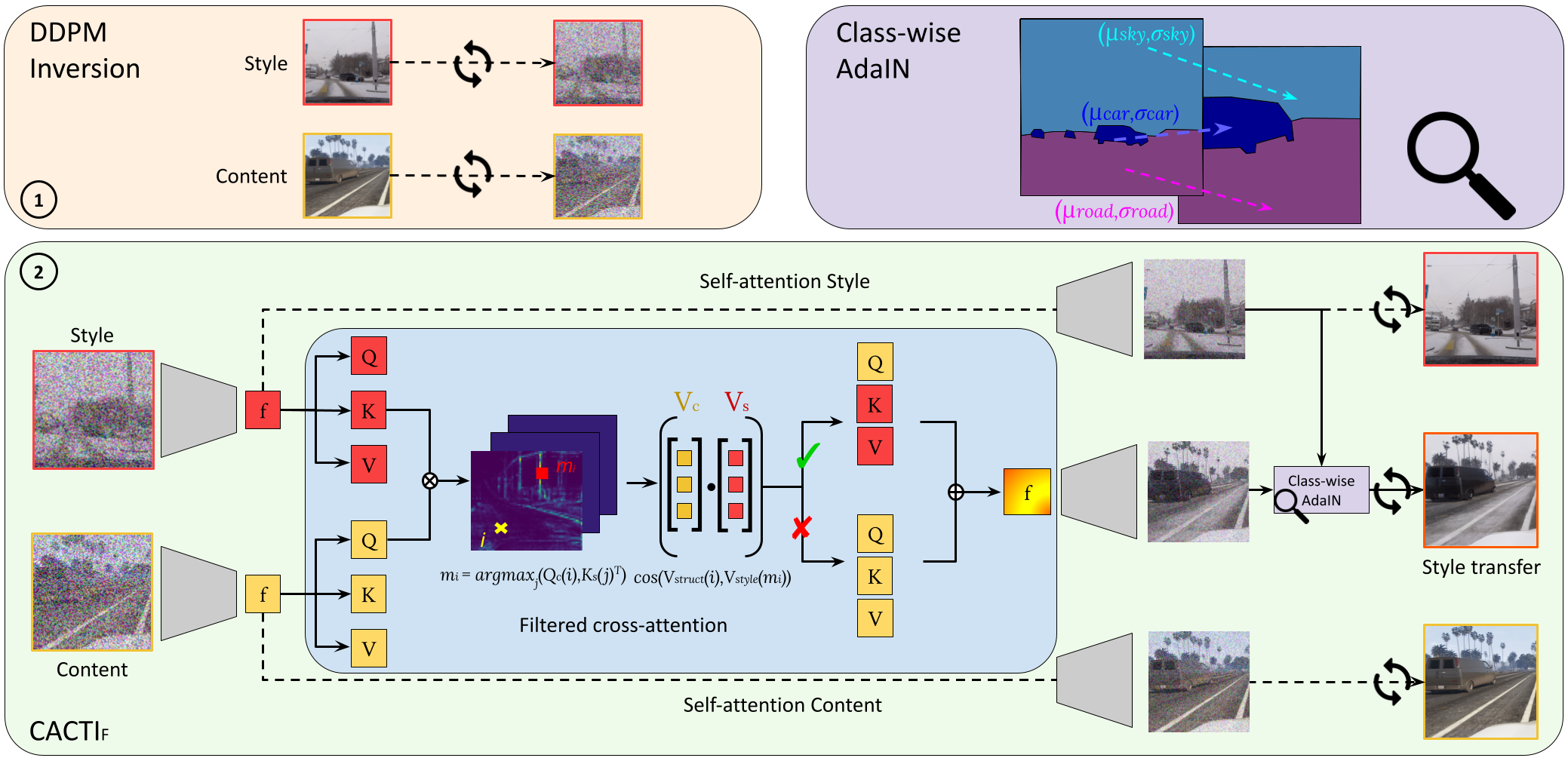}
     \caption{Overview of our proposed method for style transfer with diffusion models. 
    Given a content image and a style image, we first perform DDPM inversion (left) to obtain their respective latent representations. 
    These representations are then processed through our cross-attention mechanism, where our semantic filtering approach (middle) determines when to apply cross-image attention.
    When the KV injection yields coherent feature correspondences, cross-image attention transfers style features from the style image to the content image;
    otherwise, the original content features are preserved. 
    Finally, our class-wise AdaIN module (right) refines the style transfer by computing class-specific statistics using segmentation masks from both source and content images.}
     \label{architecture}
 \end{figure*}

\subsection{Class-wise Adaptive Instance Normalization}

In its standard form, AdaIN computes global statistics across entire feature maps in order to align them between images. This global calculation can be problematic when dealing with domain adaptation scenarios where class-specific appearances are critical for downstream tasks such as semantic segmentation.

AdaIN is a common brick in style transfer methods. It is used as a way to match feature statistics between two images. It can be used in a supervised manner by training a feature extractor from a dataset of paired images, or in a zero-shot setting with a pre-trained auto-encoder. We use the latter, within the backbone of a latent diffusion model.

As in standard applications of AdaIN with diffusion models, at each step $t$ of the denoising process we apply the AdaIN operation between the features of the style image $z_t^{style}$ and transferred image $z_t^{out}$:
\begin{equation}
    z_t^{out}\leftarrow \textrm{AdaIN}(z_t^{out}, z_t^{style}).
\end{equation}

This operation makes the features statistics from $z_t^{out}$ match those of $z_t^{style}$, effectively transferring global color distributions and contrasts. Specifically, the AdaIN operation is defined as:
\begin{equation}
\textrm{AdaIN}(x, y) = \sigma(y) \left(\frac{x - \mu(x)}{\sigma(x)}\right) + \mu(y),
\end{equation}
where $\mu(x)$ and $\sigma(x)$ denote the mean and standard deviation computed across spatial dimensions for each channel. Visually, this ensures a transfer of texture and color information. 

In our proposed \cacti\ method, we leverage the semantic segmentation masks available for both the synthetic content images and real style images to perform a class-specific statistical alignment. Rather than computing global statistics, we compute and apply statistics separately for each semantic class. This approach prevents dominant classes in the style image (such as sky or vegetation) from inappropriately influencing the appearance of other classes.

To implement \cacti, we first resize the segmentation masks to match the latent dimensions of Stable Diffusion's VAE. Then, for each semantic class $c$ present in both images, we compute:
\begin{equation}
\mu_c(z) = \frac{\sum_{i,j} M_c(i,j) \cdot z(i,j)}{\sum_{i,j} M_c(i,j)},
\end{equation}
\begin{equation}
\sigma_c(z) = \sqrt{\frac{\sum_{i,j} M_c(i,j) \cdot (z(i,j) - \mu_c(z))^2}{\sum_{i,j} M_c(i,j)}},
\end{equation}
where $M_c$ is a binary mask indicating pixels belonging to class $c$ and $(i,j)$ are spatial coordinates. We then apply the class-specific AdaIN operation:
\begin{equation}
z_t^{out}(i,j) \leftarrow \sigma_c(z_t^{style}) \left(\frac{z_t^{out}(i,j) - \mu_c(z_t^{out})}{\sigma_c(z_t^{out})}\right) + \mu_c(z_t^{style}),
\end{equation}
for all positions $(i,j)$ where $M_c(i,j) = 1$.

This class-specific approach ensures that statistical features from the style image influence only corresponding semantic regions in the content image. For example, the appearance of a "road" class in the style image affects only road pixels in the output image, preserving more realistic color distributions and reducing cross-class artifacts that can occur with global AdaIN. This is particularly important for domain adaptation scenarios where semantic consistency is crucial for downstream tasks.

\subsection{Selective Attention Filtering}

Several methods use a mechanism of cross-attention, or \textit{KV injection} for style transfer, as a means of mixing feature information from a content and style image together \cite{alaluf2024cross,deng2024z,chung_style_2024},
leveraging the self-attention layers of the denoising UNet.
When fed to a self-attention layer, a feature map $F$ is linearly projected into query $Q$, key $K$ and value $V$. Those projections, of dimension $d$ will enable to learn the interactions between different spatial positions of the feature maps.
The standard self-attention operation processes these $Q$, $K$ and $V$ projections as follows:
\begin{equation}
    \textrm{Attention}(Q,K,V) = \textrm{softmax} \left(\frac{Q\cdot K^T}{\sqrt{d}}\right) \cdot V,
\end{equation}
where the $\textrm{softmax}$ is applied independently on all rows of the matrix $Q \cdot K^T$.
In cross-image attention \cite{alaluf2024cross}, on which we base our method, the attention is computed using the content query $Q_c$ and the style key $K_s$ and value $V_s$, hence the term \textit{KV injection}. Their hypothesis is that $Q$ encapsulates the content within an image, while $K$ and $V$ represent the style information.

The style transfer of cross-image attention therefore modifies the standard attention calculation as such:
\begin{equation}
    \textrm{Attention}(Q_c,K_s,V_s) = \textrm{softmax} \left(\frac{Q_c\cdot K_s^T}{\sqrt{d}}\right) \cdot V_s.
    \label{attention}
\end{equation}

Cross-image attention establishes correspondences between semantically similar regions of source and target images, regardless of their spatial positions. 
However, in the context of synthetic-to-real domain adaptation, this approach faces a significant challenge. 
The cross-attention maps can be diffuse, with a single query in the source image attending to multiple semantically similar regions in the target image. 
This diffusion effect can lead to inconsistent style transfer and diminished adaptation performance.


To address this limitation, we propose cross-attention filtering, \cactif, as an extension of \cacti. This filtering refines the cross-image attention mechanism by keeping only the style features whose value vectors are similar to the one at the maximum attention position, reducing noise and ensuring more consistent and stable style transfer.


In the cross-attention mechanism, at each diffusion step and self-attention layer, the feature map is projected into queries ($Q$), keys ($K$), and values ($V$). The cross-attention maps $\mathcal{A}_{c,s} \in\mathbb{R}^{(h\times w)^2}$ are computed as:
\begin{equation}
    \mathcal{A}_{c,s} = Q_c \cdot K_s^T,
\end{equation}

where $Q_c$ represents queries from the content image, $K_s$ keys from the style image, $h$ and $w$ being the height and width of the self-attention layer. These attention maps indicate, for each position in the content features, which regions in the style features are most relevant.

In our proposed \cactif\ method, we selectively filter these attention maps based on feature similarity. For each position $i$ in the content image, we first identify the position $m_i$ in the style image that receives maximum attention:
\begin{equation}
    m_i = \arg\max_j(Q_c(i) \cdot K_s(j)^T).
\end{equation}

After identifying these maximum-attention correspondences, we evaluate whether the style transfer at each position is likely to produce coherent results. We compute the cosine similarity between the value vectors at corresponding positions:
\begin{equation}
s(i) = \cos\left(V_{\textrm{c}}(i), V_{\textrm{s}}(m_i)\right) = \frac{V_{\textrm{s}}(i) \cdot V_{\textrm{s}}(m_i)}{\|V_{\textrm{c}}(i)\| \|V_{\textrm{s}}(m_i)\|},
\end{equation}
where $V_{\textrm{c}}(i)$ is the value vector at position $i$ in the content image and $V_{\textrm{s}}(m_i)$ is the value vector at the corresponding position of maximum attention in the style image.

We then set a threshold $\tau$ based on a percentile $p$ of the similarity distribution:
\begin{equation}
\tau = \textrm{Percentile}(\{s(i)\}, p).
\end{equation}

For positions where the similarity falls below this threshold, indicating potential inconsistencies in the transfer, we retain the original content features rather than applying style transfer:
\begin{align}
\mathcal{A}_{\textrm{out}}(i) &= \begin{cases}
\mathcal{A}_{c, s}(i) & \text{if } s(i) \geq \tau \\
\mathcal{A}_{\textrm{c}}(i) & \text{if } s(i) < \tau
\end{cases} , \\
V_{\textrm{out}}(i) &= \begin{cases}
V_{\textrm{s}}(i) & \text{if } s(i) \geq \tau \\
V_{\textrm{c}}(i) & \text{if } s(i) < \tau
\end{cases}.
\end{align}

This selective approach preserves content structure in regions where direct style transfer might produce artifacts. The parameter $p$ controls the strictness of the filtering, with higher values resulting in more conservative style transfer. Through empirical testing, we determined that values of $p$ between 0.1 and 0.3 provide a good balance between style transfer effectiveness and content preservation.

\subsection{Combined Approach}

Our complete methodology, \cactif, combines both class-wise AdaIN and attention filtering in a complementary fashion. The class-wise AdaIN ensures appropriate appearance transfer within semantic regions, while attention filtering prevents artifacts in areas where the cross-attention correspondence between content and style is weak.

In practice, we implement our approach within the Stable Diffusion framework, applying these techniques during the denoising process.  As shown in Figure \ref{architecture}, at each diffusion step:

\begin{itemize}
\item We compute the cross-attention maps between content queries and style keys;
\item We apply attention filtering based on value vector similarity;
\item We perform the modified cross-attention operation;
\item We apply class-wise AdaIN to refine feature statistics.
\end{itemize}


We empirically find that applying these operations at multiple resolution levels in the UNet decoder (32×32 and 64×64 feature maps) yields the best results. For the percentile threshold in attention filtering, we use $p=0.25$ as our default setting based on preliminary experiments.

To evaluate the effectiveness of our proposed techniques, we conduct two complementary studies. First, we perform a qualitative analysis of the style transfer results, examining how \cacti\ and \cactif\ reduce artifacts and improve visual coherence compared to existing methods. This analysis helps isolate the specific contributions of class-wise AdaIN and attention filtering to the overall quality of the generated images.

Second, we conduct a quantitative evaluation in the context of synthetic-to-real domain adaptation for semantic segmentation. By testing our generated images as training data for segmentation models under various adverse weather conditions, we demonstrate the practical utility of our approach for downstream computer vision tasks. This evaluation framework allows us to assess not only the visual quality of our style transfer but also its effectiveness in bridging the domain gap between synthetic and real data distributions.

Together, these two studies highlighting how class-based AdaIN improves color consistency and class-specific appearance transfer, while attention filtering preserves structural integrity and reduces artifacts. The following sections detail our experimental setup and results for both evaluations.


\section{Content-preserving style transfer}
\label{sec:style}

The quality of synthetically generated images is crucial for effective domain adaptation in semantic segmentation tasks. In this section, we evaluate how our proposed methods—\cacti\  and \cactif—improve image quality compared to existing approaches. We specifically focus on reducing diffusion artifacts and enhancing alignment between segmentation masks and the corresponding generated images, which is essential for maintaining semantic consistency when transferring style from real to synthetic domains.

\subsection{Datasets}

We evaluate our methods using widely adopted datasets in the domain adaptation literature, focusing on driving scenes with semantic segmentation annotations.

\paragraph{GTA5} \cite{richter2016gta} serves as our synthetic source domain. This dataset comprises 24,966 densely annotated images rendered from the Grand Theft Auto V video game environment, with a resolution of 1914 $\times$ 1052. The synthetic nature of this dataset provides perfect ground truth labels but exhibits the characteristic domain gap when used to train models for real-world deployment.

\paragraph{Cityscapes} \cite{cordts2016Cityscapes} represents our primary real-world target domain. It consists of 5,000 densely labeled images captured in 50 European cities during daytime and under good weather conditions. These images have a resolution of 2048 $\times$ 1024 and feature typical urban driving scenarios.

\paragraph{ACDC} \cite{sakaridis2021acdc} provides our adverse condition target domains. This dataset contains 4,006 densely annotated images from driving scenarios in Switzerland, specifically focused on challenging weather conditions. The images are equally distributed across four adverse conditions: fog, nighttime, rain, and snow. Each image has a resolution of 1912 $\times$ 1024.

In our experimental framework, we address the synthetic-to-real domain adaptation scenario, using GTA5 as the source domain and either Cityscapes or ACDC as the target domain. For style transfer, we adopt a few-shot setting where we assume access to a single representative image from each target domain condition, which aligns with realistic constraints in practical applications where target domain data may be scarce.

\subsection{Experimental Setup}

We compare our proposed methods against several state-of-the-art approaches for style transfer and domain adaptation:

\paragraph{AdaIN diffusion} represents the baseline approach where standard Adaptive Instance Normalization is applied at each step of the diffusion process without class-specific modifications or cross-attention mechanisms.

\paragraph{Cross-Image Attention} implements the method from \cite{alaluf2024cross}, which uses cross-image attention with standard AdaIN to transfer appearance between images while preserving content structure.

\paragraph{DATUM} \cite{benigmim2023datum} employs personalized diffusion models for one-shot unsupervised domain adaptation, fine-tuning a text-to-image diffusion model on a single target sample to generate target-like images.

\paragraph{DGInStyle} \cite{jia2024dginstyle} generates images in the target domain style without a content reference image, operating primarily as a generative model rather than a style transfer approach.

\paragraph{\cacti} (our method) enhances Cross-Image Attention with class-specific AdaIN, leveraging segmentation labels to guide the feature statistic matching during the diffusion process.

\paragraph{\cactif} (our method) extends \cacti\ by adding selective attention filtering based on feature similarity, further reducing artifacts in regions where direct style transfer might produce inconsistencies.

For all methods based on cross-image attention (Cross-Image Attention, \cacti, and \cactif), we generate images at 512 $\times$ 1024 resolution using Stable Diffusion v1.5 with an empty prompt. To balance generation quality with computational efficiency, we skip 30 out of 50 denoising steps. For AdaIN-based methods, we apply the normalization at every step throughout the diffusion process.

While we first measure the capacity of these methods to transfer styles, we note that DATUM and DGInStyle do not fully preserve content through segmentation mask guidance. These methods usually generate images showing artifacts and undefined objects. By using style transfer and not image generation without prior, we manage to keep strong structural coherence.

\subsection{Evaluation Metrics}

We first employ two complementary metrics to evaluate the quality of generated images, studying the style transfer capacity of the proposed methods.

The Fréchet Inception Distance (FID) \cite{heusel2017fid} measures the statistical similarity between generated images and real target domain images. Lower FID values indicate better alignment with the target domain distribution, reflecting successful style transfer. For FID computation, we generate 5,000 images per method and compare against the Cityscapes dataset.

The Learned Perceptual Image Patch Similarity (LPIPS) \cite{zhang2018lpips} quantifies the perceptual distance between pairs of images, providing a measure of content preservation. We compute LPIPS between each synthetic GTA5 image and its corresponding generated version. Lower LPIPS values indicate better preservation of the original content structure.


\subsection{Results and Analysis}

\begin{figure}
\centering
\captionsetup{subrefformat=parens}
\begin{subcaptionblock}[c]{0.20\columnwidth}
    \centering
    \includegraphics[width=0.9\linewidth]{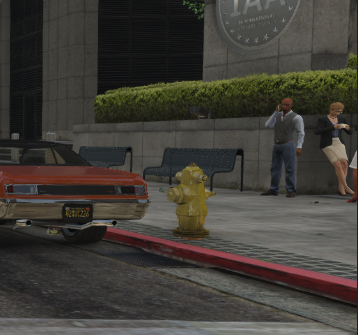}
    \caption{Content (GTA5)}
    \label{gta_13}
\end{subcaptionblock}
\begin{subcaptionblock}[c]{0.20\columnwidth}
    \centering
    \includegraphics[width=0.9\linewidth]{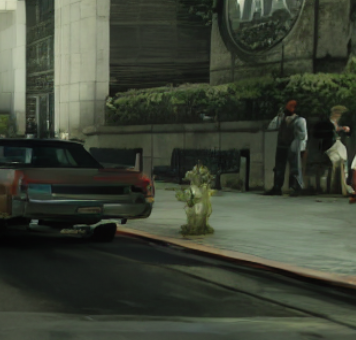}
    \caption{\cactif}
    \label{custom_13}
\end{subcaptionblock}
\begin{subcaptionblock}[c]{0.20\columnwidth}
    \centering
    \includegraphics[width=0.9\linewidth, trim=0cm 1cm 0cm 0cm, clip]{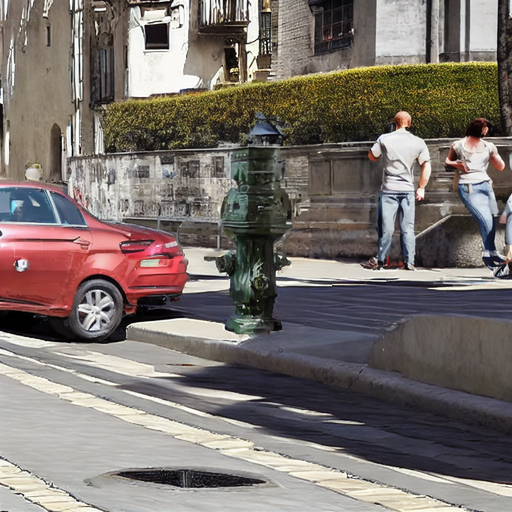}
    \caption{DGInStyle}
    \label{dgin_13}
\end{subcaptionblock}
\begin{subcaptionblock}[c]{0.20\columnwidth}
    \centering
    \includegraphics[width=0.9\linewidth, trim=0cm 1cm 0cm 0cm, clip]{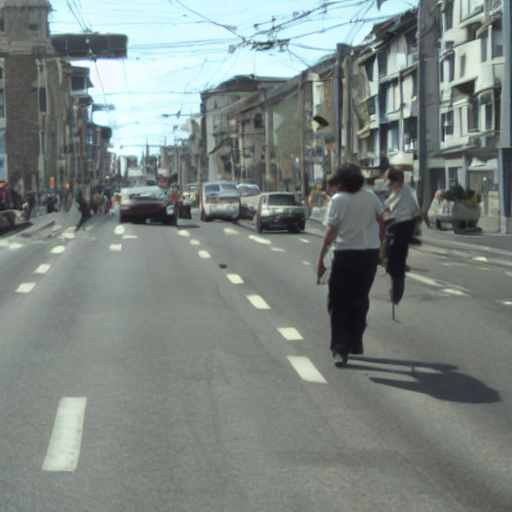}
    \caption{DATUM}
    \label{datum_city}
\end{subcaptionblock}
\begin{subcaptionblock}[c]{0.20\columnwidth}
    \centering
    \includegraphics[width=0.9\linewidth]{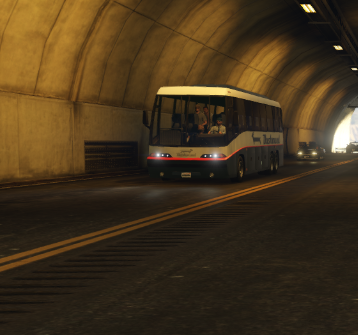}
    \caption{Content (GTA5)}
    \label{gta_49}
\end{subcaptionblock}
\begin{subcaptionblock}[c]{0.20\columnwidth}
    \centering
    \includegraphics[width=0.9\linewidth]{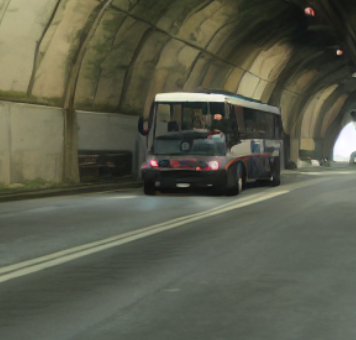}
    \caption{\cactif}
    \label{custom_49}
\end{subcaptionblock}
\begin{subcaptionblock}[c]{0.20\columnwidth}
    \centering
    \includegraphics[width=0.9\linewidth, trim=0cm 1cm 0cm 0cm, clip]{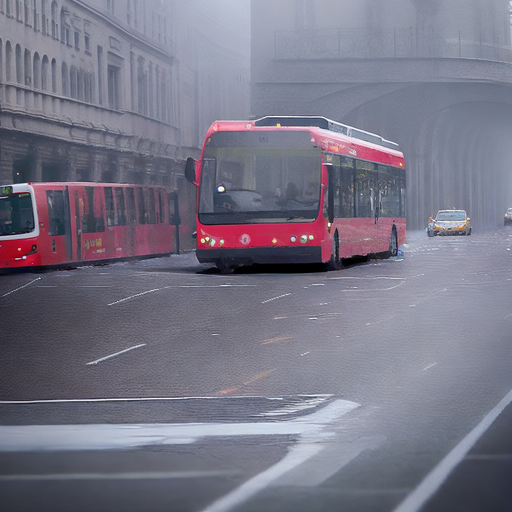}
    \caption{DGInStyle}
    \label{dgin_49}
\end{subcaptionblock}
\begin{subcaptionblock}[c]{0.20\columnwidth}
    \centering
    \includegraphics[width=0.9\linewidth, trim=0cm 1cm 0cm 0cm, clip]{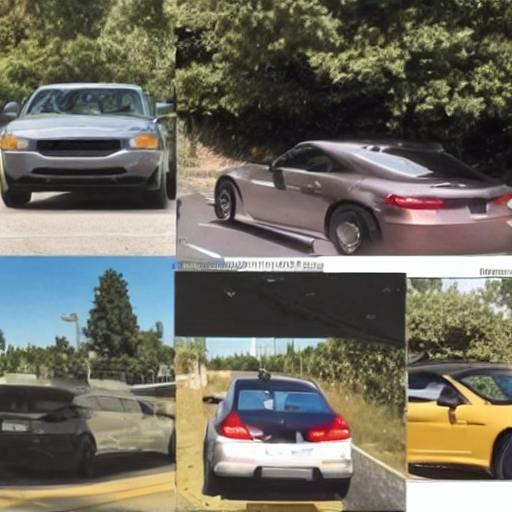}
    \caption{DATUM}
    \label{datum_night}
\end{subcaptionblock}
\caption{Qualitative comparison between dataset generation methods, with style transfer (ours) or generation without base image. Using the GTA5 dataset as base allows to keep object consistency and image structure. Note that DATUM does not rely on semantic segmentation masks and can therefore generate an image with multiple scenes, as shown here.}
\label{comparison_methods}
\end{figure}

Figure~\ref{comparison_methods} provides a visual comparison between different methods. The limitations of approaches that do not use a content reference image become immediately apparent. Methods like DGInStyle and DATUM exhibit notable artifacts and poor spatial coherence, with objects appearing distorted (as seen with the red car in Figure \ref{dgin_13}) or showing unrealistic spatial organization (Figure \ref{datum_night}).

In contrast, our proposed methods maintain object localization and structural integrity while successfully transferring the target domain style. This preservation of spatial relationships is crucial for downstream semantic segmentation tasks, as it ensures alignment between generated images and their corresponding labels.


\begin{table}
    \centering
    \begin{tabular}{| c | c | c |}
    \hline
         &  $\textrm{FID}_\textrm{cityscapes}$ $\downarrow$ & $\textrm{LPIPS}_\textrm{GTA5}$ $\downarrow$ \\
    \hline
         AdaIN diffusion & 73.26 & \textbf{0.30} \\
    \hline
         Cross-image attention & 65.31 & 0.41\\
    \hline
         DATUM & 113.12 & - \\
    \hline
         DGInStyle & 86.94 & - \\
    \hline
         \cacti  & \textbf{54.56} & 0.39 \\
    \hline
         \cactif  & 55.30 & 0.36 \\
    \hline
    \end{tabular}
    \caption{Quantitative evaluation of style transfer quality. 
$\textrm{FID}_\textrm{cityscapes}$ measures similarity to the target domain (lower is better), while $\textrm{LPIPS}_\textrm{GTA5}$ measures content preservation (lower is better). 
$\textrm{LPIPS}$ is not calculated for methods that don't use content reference images.}
    \label{tab:image_quality_results}
\end{table}

The quantitative results in Table \ref{tab:image_quality_results} confirm the visual observations. Our \cacti\ method achieves the lowest FID score (54.56) compared to all other approaches, indicating superior style transfer quality and closest resemblance to the real Cityscapes distribution. The \cactif\ variant, which adds attention filtering, shows a slight increase in FID (55.30) but significantly improves content preservation as measured by LPIPS (0.36 compared to 0.39 for \cacti).

Standard Cross-Image Attention shows reasonable style transfer capabilities (FID = 65.31) but struggles with content preservation (LPIPS = 0.41). Basic AdaIN diffusion achieves the best content preservation (LPIPS = 0.30) but at the expense of style transfer quality (FID = 73.26). We note that there is no cross-image attention in this baseline, explaining the good content preservation as only AdaIN is used for transfer between the images.

We do not calculate LPIPS for methods that do not use a content reference image, DATUM and DGInStyle, as LPIPS is calculated over image pairs. However, we note that DGInStyle shows poor performance on FID (86.94), highlighting the challenge of generating domain-specific images without structural guidance.

These results demonstrate a fundamental trade-off between style transfer fidelity and content preservation. Our proposed methods, particularly \cactif, achieve a favorable balance between these objectives, generating images that both resemble the target domain style and maintain the structural information necessary for semantic segmentation.

The class-specific approach in \cacti\ prevents dominant classes in the style image (such as sky or road) from inappropriately influencing the appearance of other classes, resulting in more realistic color distributions. The selective attention mechanism in \cactif\ further enhances content preservation by identifying and filtering potentially problematic attention mappings. 

These improvements in image quality and content preservation directly impact the effectiveness of the generated datasets for domain adaptation in semantic segmentation tasks. By reducing artifacts and maintaining structural coherence, our methods produce training data that better represents the target domain while preserving the semantic information from the source domain. This balance is essential for successful domain adaptation, as demonstrated in our subsequent semantic segmentation experiments.


\section{Style transfer for domain adaptation}
\label{sec:domain_adaptation}

While the visual quality of generated images is important, the ultimate goal of our approach is to improve performance on downstream semantic segmentation tasks through effective domain adaptation.

In practical scenarios, access to target domain data is often limited, particularly for adverse weather conditions which may be difficult or dangerous to capture.
Our method addresses this constraint by generating realistic target-domain-like images using only a small number of reference style images, enabling training of robust segmentation models without extensive target domain data collection.

In Section \ref{sec:style}, we evaluated the visual quality of images generated by our proposed methods. 
Now, we focus on assessing how these generated images perform as training data for semantic segmentation models under various domain adaptation scenarios. 
This section examines whether the improvements in style transfer quality translate to better segmentation performance when adapting from synthetic to real domains, particularly under challenging adverse weather conditions.

\subsection{Experimental Setup}

We continue using the same datasets described in Section \ref{sec:style}: GTA5 as our synthetic source domain and both Cityscapes and ACDC as our real target domains. 
For the few-shot setting, we select 1 representative image from each target domain condition (normal, fog, night, rain, and snow) to serve as style references. 
This reflects a practical scenario where a limited number of target domain samples might be available.


We explore domain adaptation in a standard scenario with the Cityscapes dataset, as well as more challenging settings with adverse weather conditions as demonstrated in the ACDC dataset.

\paragraph{Network Architectures} We evaluate our generated datasets using three network configurations:

\begin{itemize}
\item \textbf{DAFormer} \cite{hoyer2022daformer} is a state-of-the-art domain adaptation framework for semantic segmentation that employs a transformer-based encoder with adaptive context-aware fusion. 
We use DAFormer with its default MiT-B5 backbone.

\item \textbf{HRDA} \cite{hoyer2022hrda} extends DAFormer with a hierarchical multi-resolution approach specifically designed to handle high-resolution input images, which can be particularly beneficial for detecting small objects and fine details in adverse conditions.

\item \textbf{Segformer} \cite{xie2021segformer} is a transformer-based segmentation model without domain adaptation components. 
We include this model to evaluate whether our generated datasets can improve performance even without explicit domain adaptation techniques.
\end{itemize}

\paragraph{Training Protocol} For all experiments, we follow a consistent training protocol.
We first generate a dataset of 250 images for each target domain condition using the corresponding style reference image.
For DAFormer and HRDA, we employ their standard training configurations, including self-training with pseudo-labels and curriculum learning strategies.
For Segformer, we train directly on the generated images without domain adaptation components.
All models are trained for 40,000 iterations with the AdamW optimizer, following learning rate and augmentation settings from their respective original implementations.
We evaluate performance on the validation sets of Cityscapes and ACDC using the mean Intersection over Union (mIoU) metric.

\paragraph{Baselines} We compare our methods against several approaches, which we enumerate here. \textbf{Source Only} is trained only on the original synthetic GTA5 dataset without any adaptation.
\textbf{Source + Style} uses the source dataset augmented with 1 style reference image from each target domain. \textbf{AdaIN Diffusion} uses images generated with standard AdaIN during the diffusion process without class-specific modifications. We also compare with images generated from the baseline methods \textbf{Cross-Image Attention} \cite{alaluf2024cross}, \textbf{DATUM} \cite{benigmim2023datum}, and \textbf{DGInStyle} \cite{jia2024dginstyle}.


\subsection{Qualitative Analysis}

\begin{figure}
\centering
\captionsetup{subrefformat=parens}
\begin{subcaptionblock}[c]{0.28\columnwidth}
    \centering
    \includegraphics[width=\linewidth]{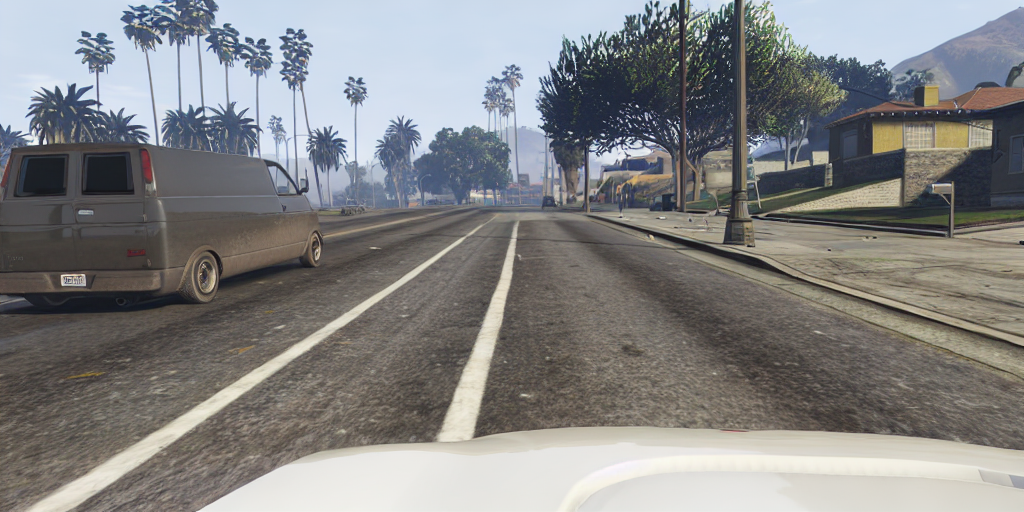}
    \caption{Content (GTA5)}
    \label{content_snow}
\end{subcaptionblock}
\begin{subcaptionblock}[c]{0.28\columnwidth}
    \centering
    \includegraphics[width=\linewidth]{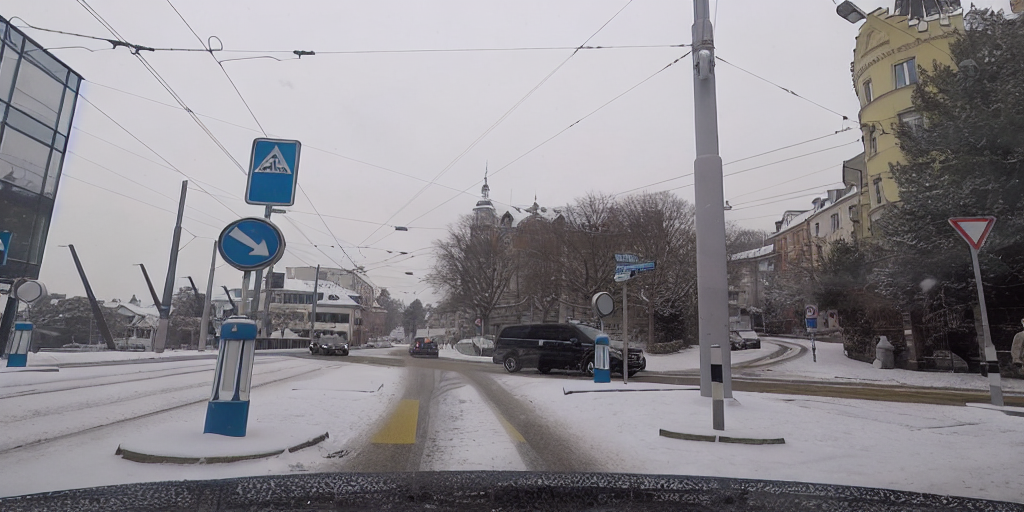}
    \caption{Style (ACDC-snow)}
    \label{style_snow}
\end{subcaptionblock}

\begin{subcaptionblock}[c]{0.28\columnwidth}
    \centering
    \includegraphics[width=\linewidth, trim=0cm 0.2cm 0cm 0.4cm, clip]{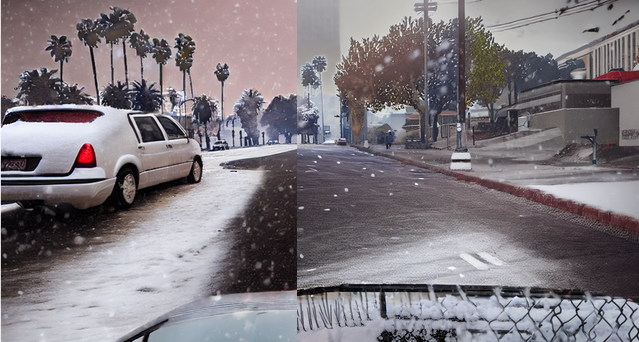}
    \caption{DGInStyle}
    \label{dginstyle_snow}
\end{subcaptionblock}
\begin{subcaptionblock}[c]{0.28\columnwidth}
    \centering
    \includegraphics[width=\linewidth]{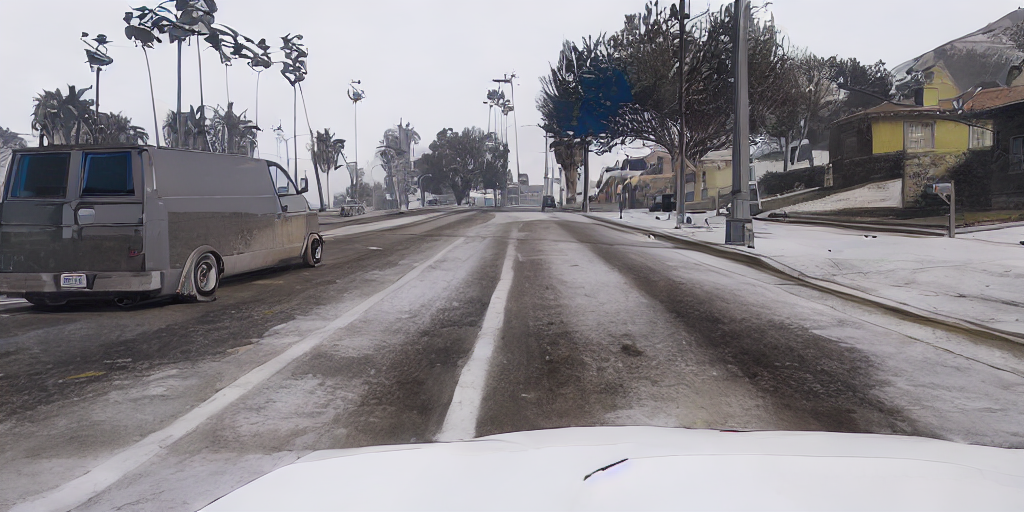}
    \caption{Cross-image attention}
    \label{cross_simple_snow}
\end{subcaptionblock}

\begin{subcaptionblock}[c]{0.28\columnwidth}
    \centering
    \includegraphics[width=\linewidth]{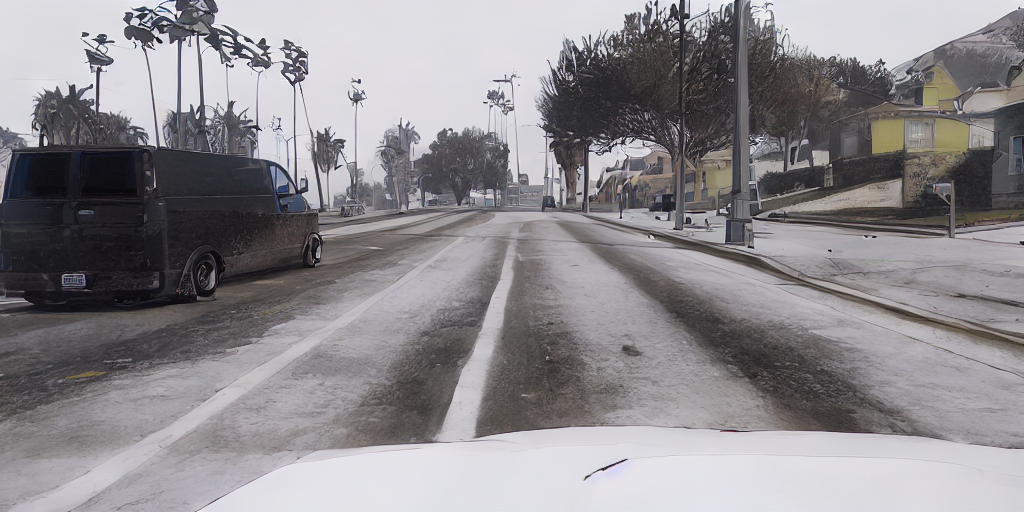}
    \caption{\cacti}
    \label{cross_class_snow}
\end{subcaptionblock}
\begin{subcaptionblock}[c]{0.28\columnwidth}
    \centering
    \includegraphics[width=\linewidth]{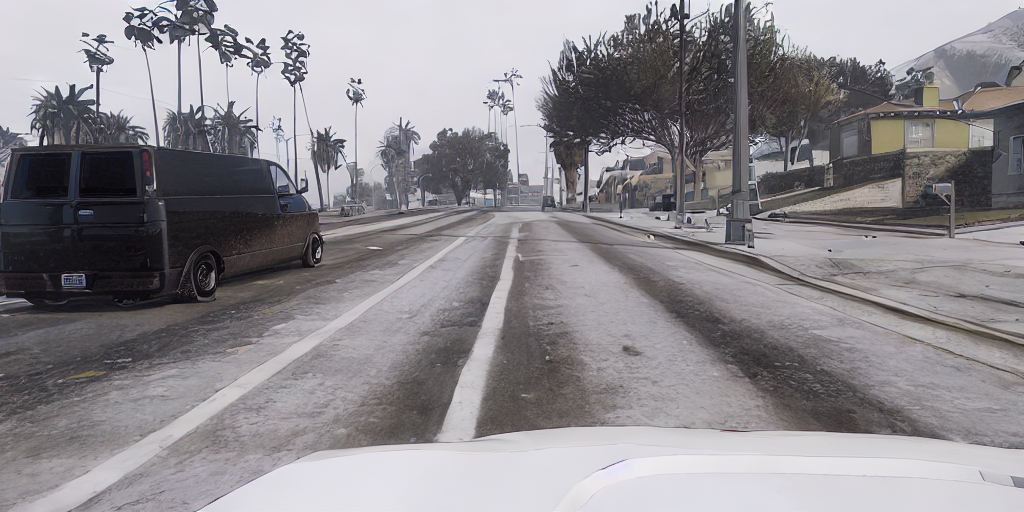}
    \caption{\cactif}
    \label{custom_class_snow}
\end{subcaptionblock}
\caption{
Style transfer comparison from GTA5 synthetic content \subref{content_snow} to ACDC-snow style \subref{style_snow}. 
The results show: DGInStyle \subref{dginstyle_snow}, Cross-Image Attention with standard AdaIN \subref{cross_simple_snow}, Cross-Image Attention with class-based AdaIN (\cacti) \subref{cross_class_snow}, and complete \cactif\ \subref{custom_class_snow}.
In dataset generation, images of 1024 by 512 pixels are used when possible. For DGInStyle, due to incompatibility with rectangular images, we show two images of 512 by 512 side-by-side.
Note the artifacts in \subref{cross_simple_snow} (blue coloration in trees), color bleeding in mountains in \subref{cross_simple_snow} and \subref{cross_class_snow}, and the more realistic snow appearance on the road in our proposed methods.
}
\label{images_snow}
\end{figure}


\begin{figure}
\centering
\captionsetup{subrefformat=parens}
\begin{subcaptionblock}[c]{0.28\columnwidth}
    \centering
    \includegraphics[width=\linewidth]{images/input/2605.png}
    \caption{Content (GTA5)}
    \label{content_night}
\end{subcaptionblock}
\begin{subcaptionblock}[c]{0.28\columnwidth}
    \centering
    \includegraphics[width=\linewidth]{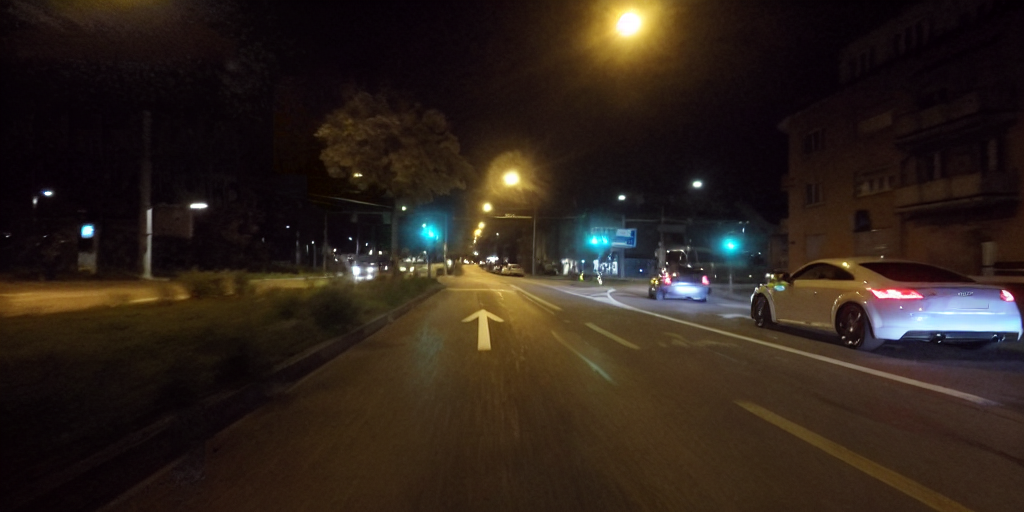}
    \caption{Style (ACDC-night)}
    \label{style_night}
\end{subcaptionblock}

\begin{subcaptionblock}[c]{0.28\columnwidth}
    \centering
    \includegraphics[width=\linewidth, trim=0cm 0.2cm 0cm 0.4cm, clip]{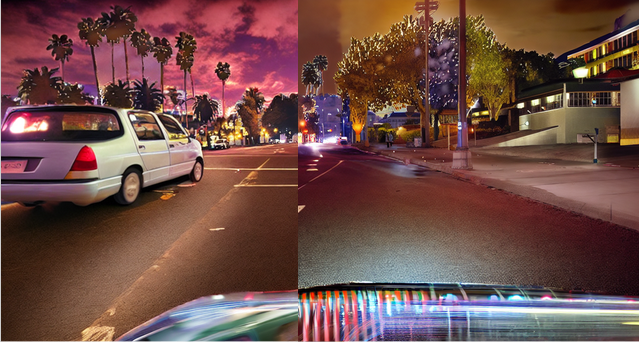}
    \caption{DGInStyle}
    \label{dginstyle_night}
\end{subcaptionblock}
\begin{subcaptionblock}[c]{0.28\columnwidth}
    \centering
    \includegraphics[width=\linewidth]{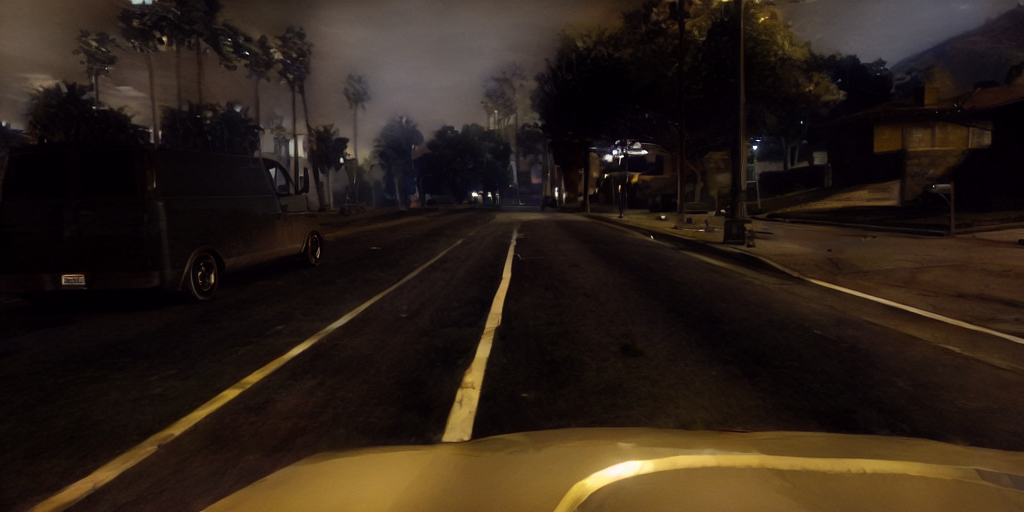}
    \caption{Cross-image attention}
    \label{cross_simple_night}
\end{subcaptionblock}

\begin{subcaptionblock}[c]{0.28\columnwidth}
    \centering
    \includegraphics[width=\linewidth]{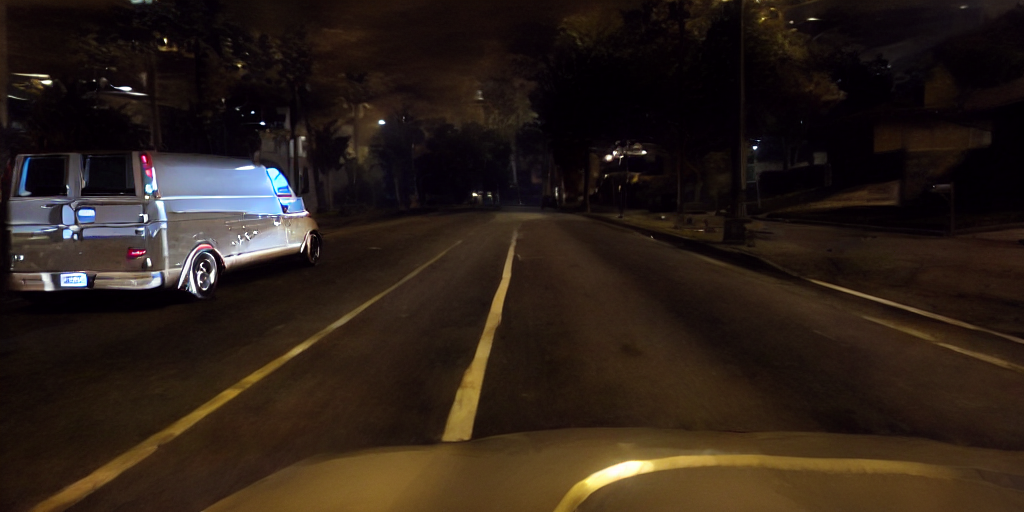}
    \caption{\cacti}
    \label{cross_class_night}
\end{subcaptionblock}
\begin{subcaptionblock}[c]{0.28\columnwidth}
    \centering
    \includegraphics[width=\linewidth]{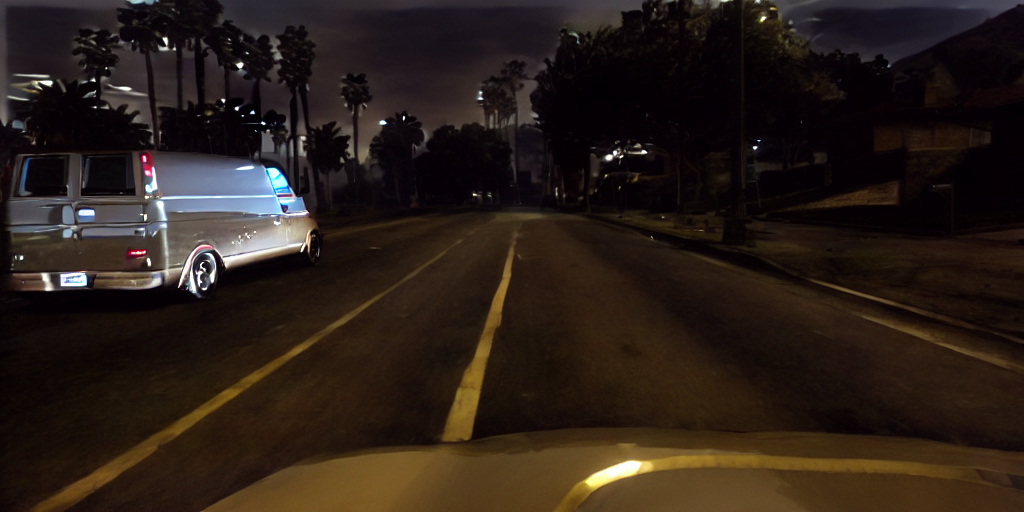}
    \caption{\cactif}
    \label{custom_class_night}
\end{subcaptionblock}
\caption{
Style transfer results from GTA5 content \subref{content_night} to ACDC-nighttime style \subref{style_night}, using the same methods as in \Cref{images_snow}.
Our methods better preserve visibility of important semantic elements while successfully transferring the nighttime appearance, maintaining better contrast and detail in critical regions.
}
\label{images_night}
\end{figure}

Figures \ref{images_snow} and \ref{images_night} provide visual comparisons of our methods when transferring styles from adverse weather conditions to synthetic images. 
In Figure \ref{images_snow}, we observe how different methods transfer the snowy appearance from an ACDC image to a synthetic GTA5 scene. 
The Cross-Image Attention approach with standard AdaIN (Figure \ref{cross_simple_snow}) introduces several artifacts, including unrealistic blue coloration in trees and yellow color bleeding into mountain regions. 
Adding class-specific AdaIN as in our \cacti\ method (Figure \ref{cross_class_snow}) partially mitigates these issues but still exhibits some color bleeding.

Our \cactif\ method (Figure \ref{custom_class_snow}) produces the most coherent results, successfully transferring the snowy appearance while maintaining the structural integrity of the original scene. 
Notably, the snow accumulation on the road surface is realistic, and the color distribution across semantic classes is consistent with real-world expectations. 
The selective attention filtering prevents inappropriate transfers between dissimilar regions, which is particularly evident when comparing the rendering of distant objects like mountains and trees.

Figure \ref{images_night} illustrates the challenge of transferring nighttime appearance, which involves significant lighting changes. 
The standard Cross-Image Attention approach (Figure \ref{cross_simple_night}) produces an image with unnatural contrasts and loss of detail in darker regions. 
Class-based AdaIN (Figure \ref{cross_class_night}) improves the results in highlighting the vehicle and darkening the sky.
Our \cactif\ method (Figure \ref{custom_class_night}) achieves a more balanced nighttime appearance, where the overall scene is appropriately darkened but important semantic elements remain discernible like the edges of the vegetation. 
This is crucial for downstream segmentation tasks, as preserving the visibility and correct appearance of traffic participants and infrastructure elements enables better model training. 
The selective attention filtering is particularly effective in this challenging lighting condition, preventing the loss of semantic information while still transferring the nighttime style characteristics.

\begin{table*}
    \centering
    \begin{tabular}{| c | c | c | c | c | c | c | c | c |}
    \hline
        & \#TS & Model & Fog & Night & Rain & Snow & Mean (ACDC) & Cityscapes \\
    \hline
    \hline
         Source + style & 5 & DAFormer & 49.90 & 17.98 & 43.22 & 39.85 & 37.91 & 53.18 \\
    \hline
         AdaIN diffusion & 5 & DAFormer & 51.01 & 16.40 & 45.72 & 43.45 & 37.88 & 50.31\\
    \hline
         Cross-image Attention & 5 & DAFormer & 51.80 & 19.37 & 45.20 & 43.80 & 40.05 & 52.36 \\
    \hline
         DATUM & 5 & DAFormer & 52.70 & 22.07 & \textbf{47.81} & \textbf{46.00} & 42.16 & \textbf{55.04}\\
    \hline
         DGInStyle & 0 & DAFormer & 49.70 & 21.06 & 44.04 & 44.28 & 40.18 & 52.42 \\
    \hline
         \cacti & 5 & DAFormer & \textbf{53.23} & \textbf{22.82} & 47.78 & 44.49 & \textbf{42.80} & 53.47 \\
    \hline
         \cactif & 5 & DAFormer & 52.88 & 21.39 & 45.93 & 43.82 & 41.13 & 54.39 \\
    \hline
    \hline
         Source + style & 5 & HRDA & 53.02 & 21.77 & 45.63 & 41.68 & 40.74 & 54.15 \\
    \hline
         \cacti & 5 & HRDA & \textbf{58.24} & \textbf{26.07} & 50.04 & \textbf{45.44} & \textbf{45.66} & 53.81\\
    \hline
         \cactif & 5 & HRDA & 58.17 & 25.61 & \textbf{51.36} & 45.11 & 45.60 & \textbf{54.82} \\
    \hline
    \hline
         Source only & 5 & Segformer & 44.32 & 16.02 & \textbf{39.03 }& 37.88 & 34.78 & 43.99\\
    \hline
         \cacti & 5 & Segformer & \textbf{45.79} & 18.23 & \textbf{39.03} & \textbf{38.88} & 36.62 & 43.30\\
    \hline
         \cactif & 5 &  Segformer & 44.53 & \textbf{21.55} & 36.79 & 38.78 & \textbf{36.71} & \textbf{48.40} \\
    \hline
    \end{tabular}
    \caption{Semantic segmentation performance (mIoU \%) on synthetic-to-real domain adaptation across different methods and models. 
We report results for each of the four adverse conditions in ACDC (Fog, Night, Rain, Snow) as well as their average (Mean ACDC) and normal conditions (Cityscapes). 
Three network architectures are evaluated: DAFormer (domain adaptation transformer), HRDA (high-resolution domain adaptation), and Segformer (without explicit domain adaptation components). 
\#TS indicates the total number of target samples used for generation. 
All results are averaged over three independent trials with different random seeds. 
Bold values indicate the best performance in each model category and condition. 
Our proposed methods (\cacti\ and \cactif) consistently outperform baselines, with \cactif\ showing particular strength when paired with the HRDA architecture.
}
    \label{tab:DA_results}
\end{table*}

\subsection{Quantitative Results}

Table \ref{tab:DA_results} presents the quantitative results for semantic segmentation performance across different models and generation methods. 
When using the DAFormer architecture, both of our proposed methods (\cacti\ and \cactif) consistently outperform the baseline approaches across most conditions. 
Specifically, \cacti\ achieves the best performance on fog and night conditions, with mIoU scores of 53.23\% and 22.82\% respectively, surpassing the Source + Style baseline by +3.33\% and +4.84\%. 
DATUM performs slightly better on rain and snow conditions, but our \cacti\ method achieves the highest mean performance across all ACDC conditions (42.80\%), demonstrating its robustness across diverse adverse conditions.

When comparing different model architectures, we observe that HRDA consistently outperforms DAFormer across all conditions when trained with both our \cacti\ and \cactif\ methods. 
The improvement is particularly significant for fog conditions, where \cacti\ with HRDA achieves 58.24\% mIoU, representing a +5.01\% improvement over the same method with DAFormer. 
This suggests that the hierarchical multi-resolution approach of HRDA is particularly effective at exploiting the high-quality and semantically consistent images generated by our method.

Examining performance across different adverse conditions reveals interesting patterns. 
The night condition remains the most challenging for all methods, with substantially lower mIoU scores than other conditions. 
However, our methods still provide considerable improvements over the baselines in this difficult scenario. 
For example, \cacti\ with HRDA achieves 26.07\% mIoU on night conditions, representing a +4.30\% improvement over the Source + Style baseline with the same architecture.

Perhaps most notably, our \cactif\ method enables even non-domain-adaptive architectures like Segformer to achieve competitive performance. 
Segformer trained with \cactif-generated images achieves 36.71\% mean mIoU across ACDC conditions and 48.40\% on Cityscapes, outperforming the Source Only baseline by +1.93\% and +4.41\% respectively. 
This demonstrates that our approach helps bridge the domain gap even without explicit domain adaptation techniques in the segmentation model.

Overall, these results confirm that our proposed methods not only generate visually appealing images but also produce semantically meaningful data that enables effective domain adaptation for semantic segmentation. 
The class-specific approach of \cacti\ ensures appropriate style transfer for different semantic regions, while the selective attention filtering in \cactif\ preserves critical structural information. 
Together, these techniques enable the generation of high-quality synthetic datasets that significantly improve segmentation performance across diverse target domains and model architectures.

\section{Conclusion and perspectives}\label{conclusion}

In this work, we proposed two novel methods for synthetic-to-real domain adaptation: Class-wise Adaptive Instance Normalization (\cacti) and Selective Attention Filtering (\cactif). 
These methods leverage diffusion models for style transfer, generating high-quality images that maintain strong structural coherence while adapting to target domain characteristics. 
Our approach specifically targets the challenging scenario of few-shot domain adaptation, where only a limited number of target domain images are available. 
The results demonstrate that class-aware style transfer can significantly improve semantic segmentation performance across various adverse weather conditions.

While our methods show clear improvements over standard style transfer approaches, the comparison with generative methods like DATUM reveals interesting trade-offs. 
DATUM achieves slightly better performance in some conditions using DAFormer, but it fundamentally transforms the content rather than preserving it. 
In contrast, our style transfer approach maintains structural coherence and semantic alignment with source images, which is critical when annotation preservation is required. 
The integration of our methods with HRDA demonstrates particularly strong results, achieving up to 58.24\% mIoU on fog conditions and a mean of 45.66\% across all adverse conditions, showing the potential of style transfer for bridging the synthetic-to-real gap.

Our experiments highlight the significant challenges that remain in domain adaptation for adverse weather conditions. 
The nighttime scenario presents the most difficulty, with our best method (\cacti\ with HRDA) achieving only 26.07\% mIoU. 
This underperformance can be attributed to the substantial appearance gap between synthetic and real nighttime scenes, as GTA5 contains fewer night examples and their visual characteristics differ significantly from real nighttime images in Cityscapes and ACDC. 
These results underscore the importance of continued research in domain adaptation techniques specifically tailored for extreme lighting and weather conditions.

Several technical aspects of our approach warrant further investigation. 
The current filtering method relies on maximum attention values, which may not capture the full complexity of attention map distributions.
Alternative pixel matching strategies, such as those employed in Eye for an Eye, could potentially improve correspondence quality, albeit at higher computational cost. 
Additionally, integrating ControlNet into the pipeline could provide more explicit guidance using segmentation masks, potentially improving structural coherence in challenging regions. 
Future work should also explore the integration of textual information through prompts to further guide the generation process.

This research demonstrates that class information can be highly valuable in guiding image generation and style transfer for domain adaptation. 
By leveraging semantic labels to constrain both statistical normalization and attention mechanisms, we can generate more realistic and semantically consistent target domain images. 
As autonomous systems continue to face deployment challenges in adverse conditions, approaches like ours that can effectively adapt models using limited target data will be increasingly important. 
Our work represents a step toward more robust perception systems that can operate reliably across diverse environmental conditions, though significant challenges remain, particularly for extreme conditions like nighttime driving.




\bibliographystyle{unsrtnat}
\bibliography{main}



\end{document}